\documentclass[11pt]{article}

\usepackage{amsmath} 
\usepackage{amssymb} 

\usepackage[utf8]{inputenc}
\usepackage{booktabs} 
\usepackage{multirow}
\usepackage{graphicx} 
\usepackage{xcolor}
\usepackage{colortbl} 

\usepackage{makecell}
\definecolor{sball}{gray}{0.95}  
\definecolor{rowgray}{gray}{0.92} 
\definecolor{graybg}{gray}{0.9}

\usepackage[dvipsnames]{xcolor}

\usepackage[preprint]{latex/acl}

\usepackage{ulem}

\usepackage{times}
\usepackage{latexsym}

\usepackage[T1]{fontenc}

\usepackage[utf8]{inputenc}

\usepackage{microtype}

\usepackage{inconsolata}

\usepackage{graphicx}

\title{LEPO: \underline{L}atent R\underline{e}asoning \underline{P}olicy \underline{O}ptimization for Large Language~Models}

\author{\textbf{Yuyan Zhou}$^{1,\dagger}$\quad
  \textbf{Jiarui Yu}$^{1,\dagger}$\quad
  \textbf{Hande Dong}$^{1}$\thanks{Corresponding Authors}\quad
  \textbf{Zhezheng Hao}$^{1,2}$\quad
  \\[3pt]
   \textbf{Hong Wang}$^{1}$\quad
  \textbf{Jianqing Zhang}$^{3}$\quad
  \textbf{Qiang Lin}$^{1}$\quad\\[3pt]
  $^{1}$ Tencent \quad 
  $^{2}$ Zhejiang University \\
  $^{3}$ Shanghai Jiao Tong University\\
  \small Emails: \url{zhou_yuyan@foxmail.com}, 
  \small\url{yujiarui9910@gmail.com},
  \small\url{donghd66@gmail.com}
}

\begin{document}
\maketitle
\begin{abstract}
\begingroup
  \renewcommand{\thefootnote}{}  
  \footnotetext{\textsuperscript{\dag} Equal Contribution.}
\endgroup
Recently, latent reasoning has been introduced into large language models (LLMs) to leverage rich information within a continuous space.
However, without stochastic sampling, these methods inevitably collapse to deterministic inference, failing to discover diverse reasoning paths.
To bridge the gap, we inject controllable stochasticity into latent reasoning via Gumbel-Softmax, restoring LLMs' exploratory capacity and enhancing their compatibility with Reinforcement Learning (RL).
Building on this, we propose \textbf{\underline{L}}atent R\textbf{\underline{e}}asoning \textbf{\underline{P}}olicy \textbf{\underline{O}}ptimization~(\textbf{LEPO}), a novel framework that applies RL directly to continuous latent representations.
Specifically, in rollout stage, LEPO maintains stochasticity to enable diverse trajectory sampling, while in optimization stage, LEPO constructs a unified gradient estimation for both latent representations and discrete tokens.
Extensive experiments show that LEPO significantly outperforms existing RL methods for discrete and latent reasoning.\footnote{Code is available at \url{https://github.com/YuyanZhou/lepo}.}
\end{abstract}

\section{Introduction}

Large Language Models (LLMs) have demonstrated remarkable reasoning capabilities via Chain-of-Thought (CoT) \citep{wei2022chain,comanici2025gemini,deepseekr1}.
However, CoT is restricted to selecting one discrete token at each reasoning step, rather than retaining the distributional information inherent in the model's internal states, which potentially induces information loss.
To address this, recent research has shifted towards latent reasoning \citep{coconut,zhang2025soft}. 
By utilizing continuous representations—such as hidden states \citep{coconut} or weighted vocabulary embeddings \citep{zhang2025soft}—this paradigm preserves richer information for more comprehensive reasoning \citep{zhu2025reasoning}.


However, due to the lack of stochastic sampling, existing latent reasoning methods \citep{coconut,zhang2025soft} yield an invariant output for a specific input. 
This deterministic process restricts the model's exploration capabilities, potentially limiting its application.
To better understand this restriction, we conceptually divide the model's exploration capabilities into two levels:
(1) \textbf{trajectory-level}: facilitating the sampling of diverse trajectories to empower the exploration of correct reasoning paths;
(2) \textbf{step-level}: performing implicit search over multiple options within each step, thereby preventing early mistakes by maintaining diverse latent possibilities.
Crucially, while current latent methods provide the step-level search through continuous representations, they sacrifice the trajectory-level exploration found in discrete approaches due to the lack of stochastic sampling.

To restore trajectory-level exploration, it is necessary to introduce stochasticity into latent reasoning.
Technically, we achieve this by adding random perturbations to the continuous representation via Gumbel-Softmax \citep{jang2016categorical,wu2025llms}.
Our motivation experiments in Section \ref{subsec: motivation} validate that simply injecting Gumbel noise effectively enhances exploration, as evidenced by increased entropy, a higher pass@32, and a shift in the problem difficulty distribution that converts intractable tasks into solvable ones.
Furthermore, this restored exploration capability inherently aligns with the exploration-exploitation nature of reinforcement learning (RL), thereby unlocking the potential for superior performance gains.

Leveraging the stochasticity provided by
Gumbel noise, 
we propose \textbf{\underline{L}}atent R\textbf{\underline{e}}asoning \textbf{\underline{P}}olicy \textbf{\underline{O}}ptimization (\textbf{LEPO}), a novel framework that applies RL directly to the latent representation.
LEPO effectively formulates latent reasoning as a learnable policy over continuous representation, allowing the model to explore and optimize latent trajectories that are essential for solving complex tasks.
During the rollout stage, we introduce controllable stochasticity via Gumbel-Softmax to sample diverse trajectories, thereby unlocking the exploration potential of latent reasoning.
In the optimization stage, we extend the optimization objective from discrete tokens to latent representations 
, enabling unified gradient estimation over both latent reasoning steps and subsequent discrete answers.
Extensive experiments show that LEPO significantly outperforms existing RL methods for discrete and latent reasoning on challenging mathematical and general-purpose benchmarks.

We summarize our major contributions as follows:
(1) We conduct motivation experiments to show the superior exploration capability of stochastic latent reasoning.
Our analysis reveals that latent reasoning with stochastic noise restores trajectory-level exploration, leading to higher entropy and facilitating more productive reinforcement learning.
(2) We propose LEPO, an RL training framework tailored for stochastic latent reasoning. By applying policy optimization directly to continuous latent representations, LEPO effectively optimizes the model's internal reasoning process.
(3) Comprehensive experiments demonstrate that LEPO outperforms baseline RL frameworks for discrete and latent reasoning on mathematical and general-purpose benchmarks.

\begin{figure*}[t]
\centering
\includegraphics[width=0.95\textwidth]{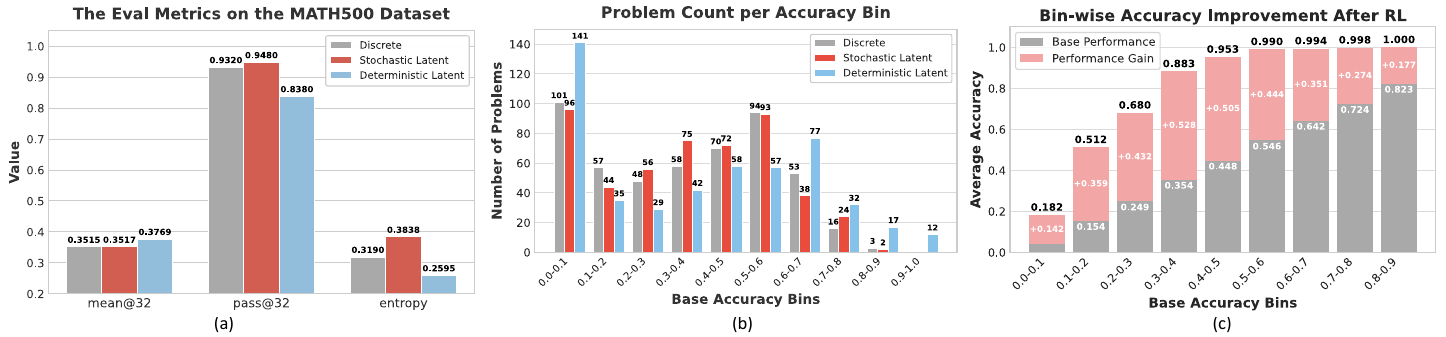}
\vspace{-3mm}
\caption{
Preliminary experiments using Qwen2.5-7B on the MATH500 dataset. 
(a) Stochastic latent reasoning enhances exploration (indicated by higher pass@32 and entropy) with comparable initial accuracy (mean@32). 
(b) Problem distribution across base accuracy bins, showing that stochastic latent reasoning shifts density from low-accuracy intervals to moderate-accuracy ones. 
(c) Performance gains after RL training, demonstrating that optimization is most effective on these moderate-accuracy problems.
}
\label{fig: motivation-fig}
\vspace{-3mm}
\end{figure*}
\section{Preliminary and Motivation} 
\label{sec: motivation}



\subsection{From Discrete to Latent Reasoning}
Standard LLMs perform reasoning via discrete token selection.
Given a query $\mathbf{q}$ and prior outputs $\mathbf{o}_{1:t-1}$, the next token $\mathbf{o}_t$ is sampled from the vocabulary distribution $\pi \in \Delta^{|V|-1}$ ,
\vspace{-2mm}
\begin{equation}
\label{eq: standard llm}
\mathbf{o}_t \sim \pi_t = \mathbf{LLM}(\mathbf{q}, \mathbf{o}_{<t}) \in \Delta^{|V|-1},
\vspace{-2mm}
\end{equation}
where $\Delta^{|V|-1}$ is the probability simplex over vocabulary $V$.
In contrast, Latent Reasoning operates on the continuous latent token, most commonly realized as the vocabulary distribution \citep{zhang2025soft}.
It defines a \textbf{Latent Token} $\mathbf{z}_t$ as:
\vspace{-2mm}
\begin{equation}
\label{eq: latent token}
\mathbf{z}_t := \pi_t = \mathbf{LLM}(\mathbf{q}, \mathbf{E}_{<t}) \in \Delta^{|V|-1},
\vspace{-2mm}
\end{equation}
where $\mathbf{E}_{<t}$ represents the input embeddings of previous steps. To proceed autoregressively, the input embedding $\mathbf{E}_t$ for the next step is calculated as the expectation of token embeddings over the latent token $\mathbf{z}_t$:
\vspace{-1mm}
\begin{equation}
\label{eq: latent input}
\mathbf{E}_t := \sum_{k=1}^{|V|} \mathbf{z}_{t,k} \mathbf{e}_k,
\vspace{-1mm}
\end{equation}
where $\mathbf{e}_k$ denote the embedding vector of the $k$-th token in the vocabulary.



\subsection{Stochasticity Really Matters} 
\label{subsec: motivation}

As shown in Eq.\ref{eq: latent token}, unlike discrete sampling, current latent reasoning methods fail to sample diverse trajectories.
\citet{wu2025llms} recently proposed using Gumbel-Softmax to restore exploratory capacity. Building upon this insight, we conduct preliminary experiments to investigate whether this stochasticity can serve as a foundation for RL.
We conduct preliminary experiments to investigate this exploration potential.
First, we observe that stochastic latent reasoning exhibits higher entropy and pass@32 compared to discrete and deterministic latent methods, indicating superior exploration capability.
Moreover, this enhanced exploration redistributes problems from the low-accuracy to the moderate-accuracy range.
This phenomenon provides more solvable yet challenging tasks for the model, thereby yielding substantial performance gains during reinforcement learning.


\textbf{Stochastic latent reasoning enhances exploration (Fig. \ref{fig: motivation-fig}a).}
Results show that stochastic latent reasoning maintains higher entropy, which naturally fosters diverse reasoning pathways, improving the discovery of correct solutions evidenced by higher pass@32.


\textbf{Exploration mitigates problem hardness (Fig. \ref{fig: motivation-fig}b).}
By categorizing problems into base accuracy bins, we observe that the problem difficulty distribution is reshaped.
Stochastic latent reasoning reduces the volume of intractable low-accuracy problems (in the 0.0-0.2 range) and converts them into solvable moderate-accuracy ones (in the 0.2-0.5 range).

\textbf{Difficulty distribution shift benefits RL optimization (Fig. \ref{fig: motivation-fig}c).}
This redistribution enriches moderate-accuracy problems, providing solvable yet challenging tasks. 
These instances offer dense reward signals, unlocking the potential for RL to achieve superior performance.

\begin{figure*}[t]
\centering
\includegraphics[width=1.0\textwidth]{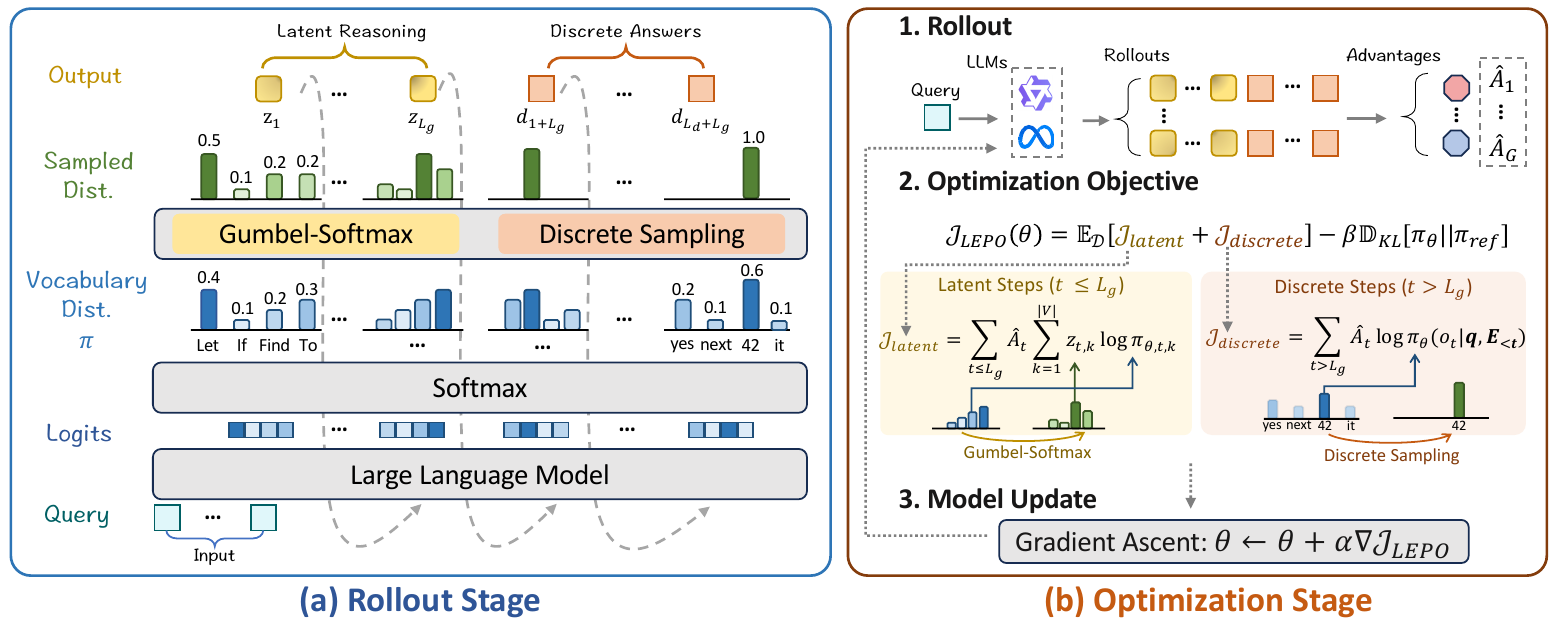}
\vspace{-7mm}
\caption{ 
Illustration of the LEPO framework. (a) Rollout Stage: We employ Gumbel-Softmax for stochastic exploration of latent tokens, followed by discrete sampling for answers. (b) Optimization Stage: We optimize a unified objective that provides consistent supervision for both latent (soft targets) and discrete (hard targets) tokens based on rollout advantages.
}
\label{fig: method-fig}
\vspace{-3mm}
\end{figure*}
\section{Methods}

Leveraging the exploration capability of stochastic latent reasoning, we design a novel RL-based framework, LEPO, for latent reasoning policy optimization.

As illustrated in Figure \ref{fig: method-fig}(a), we revitalize the rollout stage by incorporating a stochastic noise injection mechanism like Gumbel-Softmax, granting the LLM the flexibility to sample diverse trajectories.
During the LEPO rollout stage, LLMs sequentially produce latent tokens with stochastic noise for reasoning and sample discrete tokens for readable answer generation.
While in the optimization stage, beyond merely supervising discrete tokens, LEPO also integrates latent tokens into a unified gradient estimation framework.
This unified optimization provides supervision to both latent and discrete tokens, thereby enhancing its effectiveness.

\subsection{Sampling Diverse Trajectories via Gumbel-Softmax}
\label{subsec: StoLat}

Stochastic exploration is crucial in reinforcement learning. 
However, refer to Eq. \ref{eq: latent token} and \ref{eq: latent input}, without a stochastic sampling mechanism, the latent reasoning steps become deterministic given the query $\mathbf{q}$, thus hindering exploration.

To enable trajectory-level exploration, following \citet{wu2025llms}, we inject stochasticity into the continuous representation.
In our framework, this representation is embodied by the original distribution $\pi_t$ over the vocabulary.
Afterwards, we sample a stochastic latent token $\mathbf{z}_t$ derived from $\pi_t$.
The sampled $\mathbf{z}_t$ is expected to meet three criteria: 
(1) \textbf{Validity}: It needs to be a valid distribution over the vocabulary;
(2) \textbf{Latent Nature}: It is expected not to collapse into a one-hot distribution;
(3) \textbf{Faithfulness}: It needs to preserve the core information of the original distribution.

Based on these criteria, we adopt the Gumbel-Softmax to obtain the derived distribution $\mathbf{z}_t$.
Specifically, the $k$-th component of the stochastic latent token $\mathbf{z}_t$ is calculated as: 
\begin{equation}
    \mathbf{z}_{t,k} = \frac{\exp((\log \pi_{t,k} + \epsilon_k)/\tau_g)}{\sum_{j=1}^{|{V}|} \exp((\log \pi_{t,j} + \epsilon_j)/\tau_g)}, 
    \label{eq:gumbel_softmax}
\end{equation}
where $\epsilon_k$ is the independent Gumbel noise derived from $\epsilon = -\text{log}(-\text{log}(u))$ where $u \sim \mathbf{U}(0,1)$, $\tau_g$ is the temperature to balance the latent nature and faithfulness. 
As $\tau_g \rightarrow 0$, $\mathbf{z}_t$ approaches
a one-hot distribution; conversely, as $\tau_g \rightarrow \infty$, $\mathbf{z}_t$ approaches a uniform distribution.
Distinct from other noise injection strategies, Gumbel-Softmax effectively approximates multinomial sampling, thereby preserving the distributional information more faithfully \citep{wu2025llms,jang2016categorical}.

Leveraging this mechanism, we can sample diverse rollouts. Given a query $q$, we first autoregressively generate $L_g$ stochastic latent tokens. 
At each step $t$, the latent token is sampled as:
\begin{equation}
    \mathbf{z}_t = \textbf{GumbelSoftmax}(\mathbf{LLM}( \mathbf{q},\mathbf{E_{<t}}), \tau_g).
\end{equation}
Crucially, this stochastic $\mathbf{z}_t$ is then used to compute the input embedding $\mathbf{E_{t}}$ via Eq. \ref{eq: latent input} for the next step. Following the latent reasoning phase, discrete tokens $\mathbf{d}$ are sampled to generate the readable answer:
\begin{equation}
    \mathbf{d}_{t} \sim \text{LLM}(\mathbf{q}, \mathbf{E}_{1:L_g}, \mathbf{d}_{<t}),
\end{equation}
where $t=1+L_g,...,L_d+L_g$ and $L_d$ is the length of answer. This process yields a hybrid trajectory consisting of explored latent reasoning steps followed by discrete answer tokens.

\subsection{Unified Optimization Objective}
\renewcommand{\topfraction}{0.9}
\renewcommand{\dblfloatpagefraction}{0.9}

\begin{table*}[htbp]
\centering
\caption{Main results on the reasoning benchmarks. We compare \textbf{LEPO} with discrete baselines (CoT, GRPO) and prior latent reasoning methods (Soft Tokens, HRPO) across three base models: Qwen2.5 (7B, 3B) and Llama3.2-3B-Instruct. We report Pass@1 and Pass@32 accuracy on six datasets. The results demonstrate that LEPO consistently outperforms other methods, achieving superior reasoning capabilities. The best and second-best results are highlighted in \textbf{bold} and \underline{underlined}, respectively.}
\label{tab: main}

\renewcommand{\arraystretch}{0.95} 

\setlength{\tabcolsep}{1.8pt} 
\resizebox{\textwidth}{!}{%
\begin{tabular}{l cc cc cc cc cc cc >{\columncolor{sball}}c >{\columncolor{sball}}c}
\toprule[1.5pt]
\multirow{2}{*}{\textbf{Method}} & \multicolumn{2}{c}{GSM8K} & \multicolumn{2}{c}{MATH500} & \multicolumn{2}{c}{Minvervamath} & \multicolumn{2}{c}{AIME2024} & \multicolumn{2}{c}{AIME2025} & \multicolumn{2}{c}{AMC23} & \multicolumn{2}{c}{\textbf{Average}} \\
\cmidrule(r){2-3} \cmidrule(lr){4-5} \cmidrule(lr){6-7} \cmidrule(lr){8-9} \cmidrule(lr){10-11} \cmidrule(lr){12-13} \cmidrule(l){14-15}
 & Pass@1 & Pass@32 & Pass@1 & Pass@32 & Pass@1 & Pass@32 & Pass@1 & Pass@32 & Pass@1 & Pass@32 & Pass@1 & Pass@32 & \textbf{Pass@1} & \textbf{Pass@32} \\
\midrule

\multicolumn{15}{l}{\cellcolor{rowgray}\textbf{Qwen2.5-7B}} \\
\midrule
\quad CoT & 44.43 & \underline{98.48} & 35.15 & 93.20 & 14.10 & 54.04 & 4.48 & 23.33 & 3.44 & 30.00 & 25.62 & 85.00 & 21.20 & 64.01 \\
\quad CoT (w/ latent inference) & 44.21 & \textbf{98.71} & 35.17 & \textbf{94.80} & 13.57 & 55.15 & 4.90 & \textbf{36.67} & 3.65 & 30.00 & 25.16 & 85.00 & 21.11 & 66.72 \\
\quad GRPO & 89.47 & 97.57 & 74.51 & 93.40 & 33.98 & 57.72 & 10.10 & \underline{33.33} & 5.10 & 33.33 & 47.89 & 90.00 & 43.51 & \underline{67.56} \\
\quad GRPO (w/ latent inference) & \underline{89.50} & 98.03 & \underline{74.73} & 93.40 & \underline{34.03} & \underline{58.82} & 9.38 & 26.67 & 6.15 & 33.33 & 49.14 & \textbf{92.50} & \underline{43.82} & 67.13 \\
\quad Soft Tokens \citep{butt2025soft} & 87.08 & 94.39 & 73.57 & 89.80 & 30.33 & 47.06 & \underline{10.94} & 23.33 & \underline{7.29} & \textbf{36.67} & \underline{50.47} & 82.50 & 43.27 & 62.29 \\
\quad HRPO \cite{yue2025hybrid} & 85.97 & 89.46 & 66.60 & 86.60 & 22.79 & 43.38 & 10.00 & 20.00 & \underline{7.29} & 23.33 & 47.50 & 80.00 & 40.02 & 57.13 \\
\quad \textbf{LEPO (Ours)} & \textbf{90.28} & 98.03 & \textbf{75.53} & \underline{93.80} & \textbf{34.06} & \textbf{59.56} & \textbf{11.04} & \textbf{36.67} & \textbf{8.33} & \textbf{36.67} & \textbf{52.58} & \textbf{92.50} & \textbf{45.30} & \textbf{69.54} \\

\midrule
\multicolumn{15}{l}{\cellcolor{rowgray}\textbf{Qwen2.5-3B}} \\
\midrule
\quad CoT & 72.60 & \underline{98.18} & 52.38 & 90.80 & 16.61 & 50.00 & 3.23 & 20.00 & 1.98 & \textbf{30.00} & 24.22 & 80.00 & 28.50 & 61.50 \\
\quad CoT (w/ latent inference) & 72.93 & \textbf{98.48} & 52.40 & \underline{91.40} & 16.91 & \underline{50.74} & 3.33 & 23.33 & 1.98 & \underline{26.67} & 25.11 & 82.50 & 28.78 & 62.19 \\
\quad GRPO & \underline{81.80} & 97.42 & 61.00 & 90.00 & \underline{22.84} & \textbf{51.47} & 6.46 & \underline{26.67} & 2.08 & 23.33 & 33.36 & \underline{87.50} & 34.59 & \underline{62.73} \\
\quad GRPO (w/ latent inference) & 81.35 & 97.42 & \underline{62.59} & 90.60 & 21.69 & 50.37 & 6.67 & 23.33 & \underline{3.10} & \underline{26.67} & 32.74 & \underline{87.50} & \underline{34.69} & 62.65 \\
\quad Soft Tokens \citep{butt2025soft} & 77.56 & 91.74 & 55.26 & 82.60 & 17.13 & 33.82 & 4.48 & 20.00 & 1.15 & 23.33 & 27.42 & 77.50 & 30.50 & 54.83 \\
\quad HRPO \cite{yue2025hybrid} & 77.63 & 89.46 & 56.40 & 84.00 & 16.54 & 33.09 & \underline{7.50} & 23.33 & \textbf{3.33} & 20.00 & \underline{35.00} & 77.50 & 32.73 & 54.56 \\
\quad \textbf{LEPO (Ours)} & \textbf{82.94} & 97.57 & \textbf{63.23} & \textbf{91.60} & \textbf{24.22} & \underline{50.74} & \textbf{7.81} & \textbf{30.00} & \underline{3.10} & \underline{26.67} & \textbf{35.08} & \textbf{90.00} & \textbf{36.06} & \textbf{64.43} \\

\midrule
\multicolumn{15}{l}{\cellcolor{rowgray}\textbf{Llama3.2-3B}} \\
\midrule
\quad CoT & 73.22 & 96.89 & 43.75 & 84.40 & 16.22 & 45.96 & 6.77 & 26.67 & 0.52 & 13.33 & 21.48 & 75.00 & 26.99 & 57.04 \\
\quad CoT (w/ latent inference) & 73.32 & \textbf{97.42} & 43.59 & \underline{85.00} & 15.80 & 45.96 & 7.19 & 30.00 & \underline{0.83} & \textbf{16.67} & 21.64 & \textbf{77.50} & 27.06 & \underline{58.76} \\
\quad GRPO & 74.69 & \underline{97.27} & 43.89 & 83.20 & \textbf{16.67} & \underline{47.06} & 6.67 & \underline{33.33} & 0.42 & 13.33 & \underline{24.14} & 75.00 & 27.73 & 58.20 \\
\quad GRPO (w/ latent inference) & 74.41 & 96.66 & 43.86 & 84.80 & 16.74 & \underline{47.06} & 7.19 & \underline{33.33} & 0.42 & 16.67 & 23.98 & 72.50 & \underline{27.77} & 58.50 \\
\quad Soft Tokens \citep{butt2025soft} & 75.27 & 93.03 & \underline{44.08} & 81.60 & \underline{16.91} & 43.38 & 4.79 & 30.00 & 0.42 & 10.00 & 20.55 & 72.50 & 27.00 & 55.10 \\
\quad HRPO \cite{yue2025hybrid} & \underline{77.10} & 90.30 & 39.60 & 79.60 & 15.81 & 41.18 & \textbf{10.00} & 26.67 & 0.42 & 6.67 & 17.50 & 67.50 & 26.73 & 51.97 \\
\quad \textbf{LEPO (Ours)} & \textbf{77.29} & 96.97 & \textbf{46.51} & \textbf{86.20} & \textbf{17.65} & \textbf{48.16} & \underline{9.38} & \textbf{40.00} & \textbf{0.96} & \textbf{16.67} & \textbf{27.03} & \textbf{77.50} & \textbf{29.80} & \textbf{60.92} \\
\bottomrule[1.5pt]
\end{tabular}%
}
\end{table*}
Inspired by recent RL methods \citep{shao2024deepseekmath,yu2025dapo}, we introduce a unified framework for effective joint optimization of latent and discrete tokens.

Given a query $\mathbf{q}$, we first sample a group of $G$ rollouts. Based on a simple outcome-based reward $r$ (1 for correct rollout and 0 otherwise), we calculate the advantages for the $i$-th rollout using normalized rewards:
\begin{equation}
    \hat{A}_i = \frac{r_i - \text{mean}([r_1,...,r_G])}{\text{std}([r_1,...,r_G])}.
\end{equation}

We then employ an on-policy RL algorithm to estimate the gradients. The optimization objective is divided into two parts based on the reasoning stage.

\noindent\textbf{For Discrete Tokens ($t > L_g$):} The objective follows the standard REINFORCE formulation. We maximize the log-likelihood of the selected token $o_{i,t}$ as:
\begin{equation}
    \mathcal{J}_{\text{discrete}}^{(i)} = \sum_{t > L_g} \hat{A}_{i,t} \log \pi_{\theta}(o_{i,t} | \mathbf{q}, \mathbf{E}_{<t}).
\end{equation}

\noindent\textbf{For Latent Tokens ($t \le L_g$):} Since the latent token $\mathbf{z}_{i,t}$ is a continuous distributional vector rather than a discrete index, the standard probability mass function does not apply. To enable consistent supervision, we propose a soft surrogate objective. We treat the sampled stochastic vector $\mathbf{z}_{i,t}$ (from Eq. \ref{eq:gumbel_softmax}) as a "soft label" and maximize the alignment between the current policy $\pi_\theta$ and this label, weighted by the advantage:
\vspace{-2mm}
\begin{equation}
    \mathcal{J}_{\text{latent}}^{(i)} = \sum_{t \le L_g} \hat{A}_{i,t} \sum_{k=1}^{|V|} \mathbf{z}_{i,t,k} \log \pi_{\theta, k}(\mathbf{q}, \mathbf{E}_{<t}).
    \raisetag{15pt} 
\end{equation}
This term generalizes the discrete objective: if $\mathbf{z}_{i,t}$ were a one-hot vector, this equation would strictly recover the discrete policy gradient.


\noindent\textbf{Unified Objective:} Finally, let $\mathcal{J}_{\text{total}}^{(i)} = \mathcal{J}_{\text{latent}}^{(i)} + \mathcal{J}_{\text{discrete}}^{(i)}$ denote the total objective for the $i$-th trajectory. 
The overall LEPO optimization objective is computed as:
\begin{equation}
\begin{split}
    \mathcal{J}_{\text{LEPO}}(\theta) &= \mathbb{E}_{\mathcal{D}} \bigg[ \frac{1}{G} \sum_{i=1}^{G} \frac{1}{T_i} \mathcal{J}_{\text{total}}^{(i)}  \bigg] \\
    &\quad - \beta \mathbb{D}_{\text{KL}}[\pi_{\theta} || \pi_{\text{ref}}],
\end{split}
\raisetag{15pt} 
\label{eq:final_objective}
\end{equation}
where $T_i$ is the length of the $i$-th trajectory, $\pi_{\text{ref}}$ is the reference model for the KL-divergence regularization, and $\beta$ is the coefficient. 
This unified framework ensures that both the latent reasoning process and the final answer generation are optimized towards high-reward outcomes.

\section{Experiments}

\subsection{Experimental Settings}
\begin{table*}[htbp]
\centering
\caption{Out-of-distribution (OOD) evaluation on general reasoning tasks. We report Pass@1 and Pass@32 accuracy on GPQA-Diamond, ARC-C, and MMLU-STEM datasets. 
}
\label{tab: ood}
\resizebox{0.85\textwidth}{!}{%
\begin{tabular}{@{\hspace{2mm}}l cc cc cc >{\columncolor{sball}}c >{\columncolor{sball}}c}
\toprule[1.5pt]
\multirow{2}{*}{\textbf{Method}} & \multicolumn{2}{c}{GPQA-Diamond} & \multicolumn{2}{c}{ARC-C} & \multicolumn{2}{c}{MMLU-STEM} & \multicolumn{2}{c}{\textbf{Average}} \\
\cmidrule(r){2-3} \cmidrule(lr){4-5} \cmidrule(lr){6-7} \cmidrule(l){8-9}
 & Pass@1 & Pass@32 & Pass@1 & Pass@32 & Pass@1 & Pass@32 & \textbf{Pass@1} & \textbf{Pass@32} \\
\midrule

\multicolumn{9}{l}{\cellcolor{rowgray}\textbf{Qwen2.5-7B}} \\
\midrule
\quad CoT & 10.92 & 82.83 & 65.34 & \textbf{99.91} & 58.17 & \textbf{99.90} & 44.81 & 94.21 \\
\quad CoT (w/ latent inference) & 10.65 & 83.33 & 64.51 & \textbf{99.91} & 57.72 & \underline{99.84} & 44.29 & 94.36 \\
\quad GRPO & 28.28 & 87.88 & \underline{87.08} & \underline{98.98} & \underline{78.75} & 97.18 & \underline{64.70} & 94.68 \\
\quad GRPO (w/ latent inference) & 28.09 & \underline{88.89} & 87.07 & 98.72 & 78.71 & 97.37 & 64.62 & \underline{94.99} \\
\quad Soft Tokens \citep{butt2025soft} & \underline{28.69} & 81.31 & 86.67 & 95.48 & 76.63 & 93.24 & 63.99 & 90.01 \\
\quad HRPO \cite{yue2025hybrid} & 27.84 & 78.79 & 86.12 & 94.71 & 76.40 & 92.77 & 63.45 & 88.76 \\
\quad LEPO (Ours) & \textbf{29.96} & \textbf{90.91} & \textbf{87.20} & 98.89 & \textbf{79.05} & 97.40 & \textbf{65.40} & \textbf{95.73} \\

\bottomrule[1.5pt]
\end{tabular}%
}
\end{table*}


\textbf{Baselines.}
We compare LEPO against a diverse array of baselines across three paradigms. In the realm of \textit{inference-only methods}, we include CoT and CoT with latent inference; for \textit{discrete-token RL}, we select GRPO \citep{shao2024deepseekmath} for comparison; while for \textit{latent-token RL}, Soft Tokens \citep{butt2025soft} and HRPO \citep{yue2025hybrid} are employed as baselines.
Additionally, we employ Gumbel-Softmax latent inference to the model trained by GRPO as a supplement.

\textbf{Training and Evaluation Settings.}
We employ three prominent open-source models, i.e., Qwen2.5-7B, Qwen2.5-3B, and Llama-3.2-3B-Instruct as the base models to benchmark LEPO.
The three models are trained on DAPO-MATH-17k \citep{yu2025dapo} and then evaluated on GSM8k, MATH500, Minervamath, AIME2024, AIME2025, and AMC23.
To evaluate generalization and ensure capability retention, we also conduct OOD evaluation in three additional benchmarks: GPQA-Diamond \citep{rein2024gpqa}, ARC-C \citep{allenai:arc}, and MMLU-STEM \citep{hendryckstest2021}.
During the evaluation phase, we report Pass@1 and Pass@32 metrics to assess the quality of the learned policy and the coverage of the solution space, respectively.
We utilize a maximum response length of 2,048, top-p=0.95, and top-k=30. The temperature is adjusted to 1.0 for training and 0.6 for evaluation.


\subsection{Results}

\textbf{Main Results.}
We compare LEPO to all baselines and report the results in Table \ref{tab: main}. 
Compared to the discrete GRPO, LEPO improves Pass@1 by 1.76$\%$ and Pass@32 by 1.95$\%$ on average. Furthermore, compared to previous SOTA latent RL methods, Soft Tokens, it achieves gains of 3.46$\%$ and 7.56$\%$.
The training reward curves in Figure \ref{fig: ablation-fig}c show that LEPO exhibits superior convergence performance compared to GRPO.
These results demonstrate that LEPO effectively enhances the model's reasoning capabilities.

\textbf{OOD Results.}
We conducted experiments on three out-of-distribution (OOD) tasks based on Qwen2.5-7B shown in Table \ref{tab: ood}.
The results show that LEPO consistently achieves superior pass@1 performance across all OOD benchmarks, indicating its robust generalization capabilities after training.

\subsection{Training Dynamics and Analysis}
\begin{figure}
    \centering
    \includegraphics[width=0.8\linewidth]{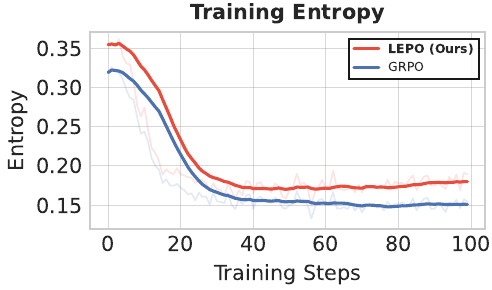}
    
    \vspace{-3mm}
    
    \caption{
    Evolution of entropy. Comparison of entropy changes between LEPO and GRPO during the training process.
    }
    \label{fig: entropy-fig}
    
    \vspace{-5mm}
\end{figure}
\begin{figure}
    \centering
    \includegraphics[width=0.8\linewidth]{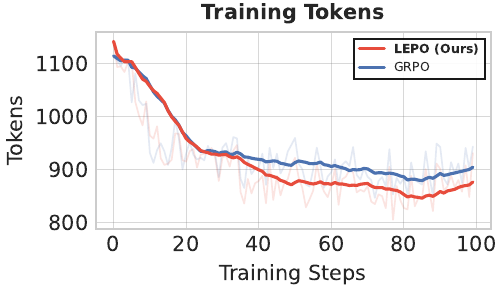}
    
    \vspace{-3mm}
    
    \caption{
    Comparison of token consumption during the training. The plot shows the change in token counts over training steps.
    }
    \label{fig: tokens-fig}
    
    \vspace{-5mm}
\end{figure}
\begin{figure*}[t]
\centering
\includegraphics[width=0.92\textwidth]{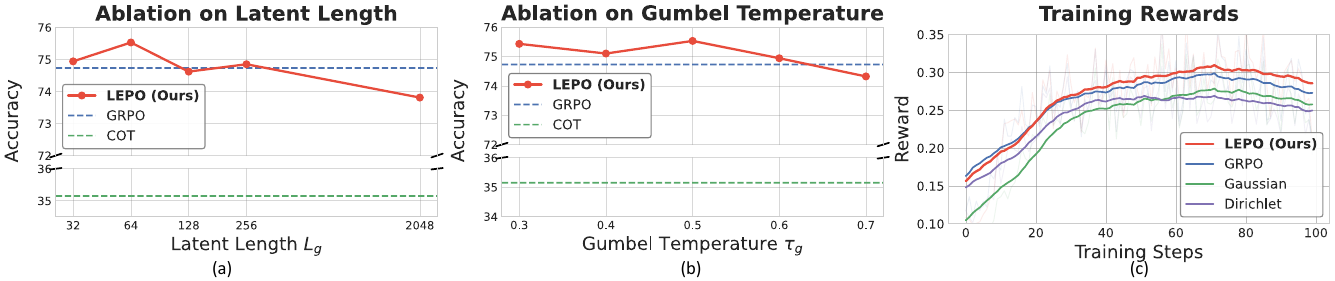}
\vspace{-3mm}
\caption{
Analysis of Hyperparameters and Training Dynamics using Qwen2.5-7B.
(a) Performance comparison on MATH500 across varying latent lengths. 
(b) Performance comparison across varying Gumbel temperatures.
(c) Training reward curves demonstrate that LEPO with Gumbel noise exhibits superior convergence compared to GRPO, as well as the Gaussian and Dirichlet noise variants.
}
\label{fig: ablation-fig}
\vspace{-2mm}
\end{figure*}

\textbf{Entropy.}
We track the evolution of entropy during the training process based on Qwen2.5-7B shown in Figure \ref{fig: entropy-fig}. 
It can be observed that the entropy exhibits a downward trend as training progresses. 
The entropy of LEPO remains consistently higher than that of GRPO, indicating that LEPO effectively maintains the exploratory nature of latent reasoning throughout training.

\textbf{Token Efficiency.}
We report the evolution of token consumption for Qwen2.5-7B in Figure \ref{fig: tokens-fig}. 
Initially, the token counts for LEPO and GRPO are comparable. 
As the model trains, LEPO encourages the utilization of information-rich continuous representations, which results in a distinct reduction in token consumption compared to the baseline.

\textbf{Latent Length $L_g$.} 
We present the results for different latent lengths using Qwen2.5-7B in Fig. \ref{fig: ablation-fig}a. 
It demonstrates that LEPO achieves superior accuracy compared to the GRPO baseline across a broad range of latent lengths,  especially with compact latent representations.

\textbf{Gumbel Temperature $\tau_g$.}
With the latent length fixed at 64, we examine how model performance varies with $\tau_g$.
The results in Fig. \ref{fig: ablation-fig}b indicate that LEPO consistently outperforms GRPO within the effective range ($\tau_g \in [0.3,0.6]$), showing robust improvements under the Gumbel sampling configurations.

\begin{figure*}[t]
\centering
\includegraphics[width=0.92\textwidth]{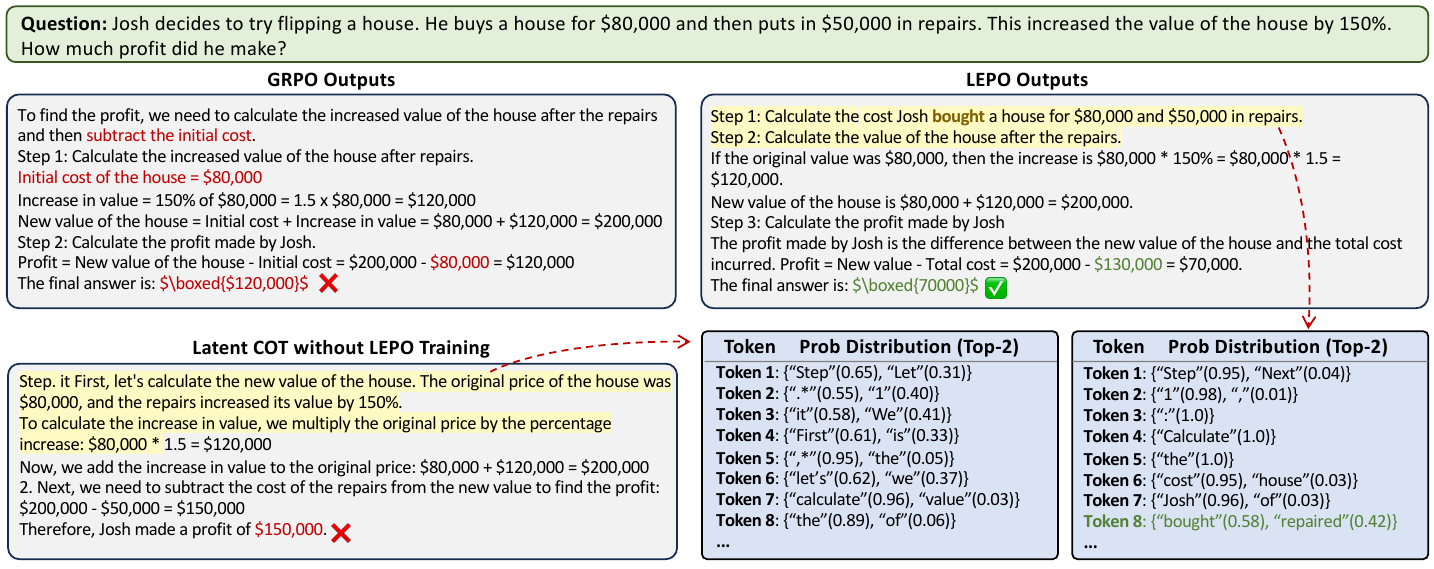}
\vspace{-1mm}
\caption{
Qualitative comparison of GRPO, LEPO and latent COT on an arithmetic task from GSM8k.
We highlight the latent tokens in yellow.
}
\label{fig: case-fig}
\vspace{-5mm}
\end{figure*}
\textbf{Noise Type.}
We modified the noise injection method to use Gaussian and Dirichlet noise. As shown in Fig. \ref{fig: ablation-fig}c, the methods with these types of noise have inferior convergence performance to that of GRPO and LEPO with Gumbel noise.

\textbf{Case Study.}
We showcase a comparison of the reasoning steps and final answers produced by latent CoT, GRPO, and LEPO on an arithmetic task from GSM8K. As illustrated in Figure \ref{fig: case-fig}, without latent reasoning, the discrete method GRPO is misled by the one-hot tokens ``initial cost'' and overlooks ``repair expense'', ultimately failing to reach the correct answer. While latent CoT leverages continuous latent tokens, without specialized training, it fails to deduct the 'original price' from the total, leading to an incorrect search path. In contrast, LEPO exhibits greedy-like behavior on most tokens by producing peaky distributions (e.g. token 4 ``Calculate''), yet maintains meaningful uncertainty at key steps (e.g. token 8 ``bought'' and ``repaired''). Consequently, LEPO achieves a superior trade-off between inference efficiency and path exploration, leading to more accurate and robust reasoning.

\section{Related Work}
Based on the training paradigms, latent reasoning methods can be categorized into three types: training-free, pre-training, and post-training.
(1) \textbf{Training-free} latent methods use the weighted sum of vocabulary embeddings for latent reasoning without training
\citep{zhang2025soft,zhuang2025text,wu2025llms,shi2025swireasoning}.
(2) \textbf{Pre-training} latent methods attempt to incorporate continuous representations for direct reasoning during the pre-training phase, commonly via interleaving continuous and discrete tokens \citep{tack2025llm} or employing internal continuous CoT \citep{geiping2025scaling, zhu2025scaling}.
(3) \textbf{Post-training} latent methods can be categorized into SFT-based and RL-based approaches. 
SFT-based methods \citep{coconut,shen2025codi,cheng2024compressed} provide implicit supervision to latent representations by aligning discrete answer tokens with target ones. 
RL-based methods sample trajectories containing latent tokens and utilize rewards for supervision
\citep{yue2025hybrid,butt2025soft,zheng2025soft,tan2025think}. 
\citet{butt2025soft} employs RLOO to optimize embeddings perturbed by Gaussian noise. 
\citet{zheng2025soft} utilizes the reparameterization trick for gradient estimation.
\citet{yue2025hybrid} mix latent and sampled discrete tokens at each step to introduce stochasticity.
Our proposed \textbf{LEPO} falls into the category of latent RL methods. 
Distinct from these approaches, LEPO provides a unified optimization objective that supports the simultaneous optimization of both latent and discrete tokens within a single trajectory. 
Furthermore, instead of relying on Gaussian noise, we employ the Gumbel-Softmax mechanism to encourage the exploration of diverse trajectories.
Empirically, our method achieves state-of-the-art performance.

\section{Conclusion}


In this work, we identify the limitation of deterministic inference in existing latent reasoning, which hinders trajectory-level exploration.
By injecting controllable stochasticity via Gumbel-Softmax, we recover the model's capability to sample diverse reasoning paths.
Building on this, we propose \textbf{LEPO}, a training framework that applies RL directly to continuous representations.
Specifically, LEPO constructs a unified gradient estimation for both latent representations and discrete tokens, enabling the efficient optimization of internal reasoning policies.


\section{Limitations}
Despite the effectiveness of LEPO in enhancing reasoning capabilities, we acknowledge two primary limitations.

First, regarding interpretability, unlike discrete Chain-of-Thought methods that produce human-readable traces, our latent reasoning steps operate in a continuous space, which is not directly intelligible to humans. However, we argue that this is an inherent trade-off for leveraging the rich, dense information within vocabulary distributions. Furthermore, LEPO mitigates this issue by restricting the continuous representation strictly to the reasoning phase; the final answer generation phase utilizes standard discrete tokens, ensuring that the outcome remains fully readable. Additionally, for diagnostic purposes, the semantic meaning of the latent reasoning process can still be approximately interpreted by decoding the top-1 token from the latent distributions.

Second, concerning the generation paradigm, the current LEPO framework adopts a sequential "latent-first, discrete-later" structure. While this design effectively disentangles exploration from expression for mathematical benchmarks, it does not currently support the dynamic interleaving of latent and discrete tokens (i.e., alternating between thinking and writing). We believe that extending LEPO to support such a hybrid, interleaved generation mode could further enhance flexibility for open-ended tasks, representing a promising direction for future work.

\bibliography{latex/references}

\appendix

\section{More Experimental Setup}

\subsection{Prompt Examples}
\begin{figure*}[t]
\centering
\includegraphics[width=1.0\textwidth]{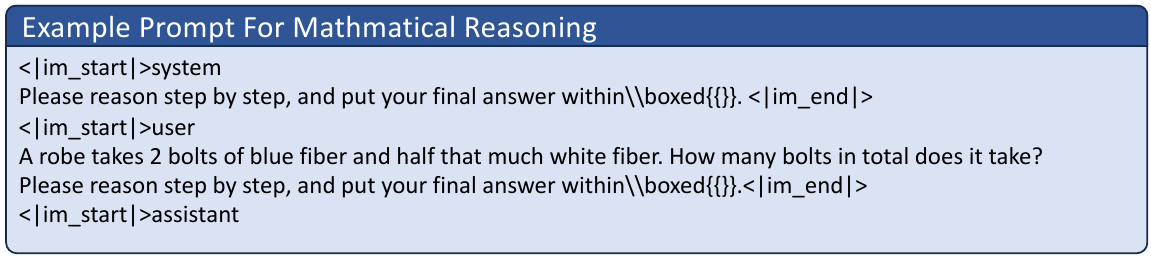}
\vspace{-7mm}
\caption{ 
Example Prompt for Mathematical Reasoning in LEPO.
}
\label{fig: prompt-fig1}
\vspace{-3mm}
\end{figure*}
\begin{figure*}[t]
\centering
\includegraphics[width=1.0\textwidth]{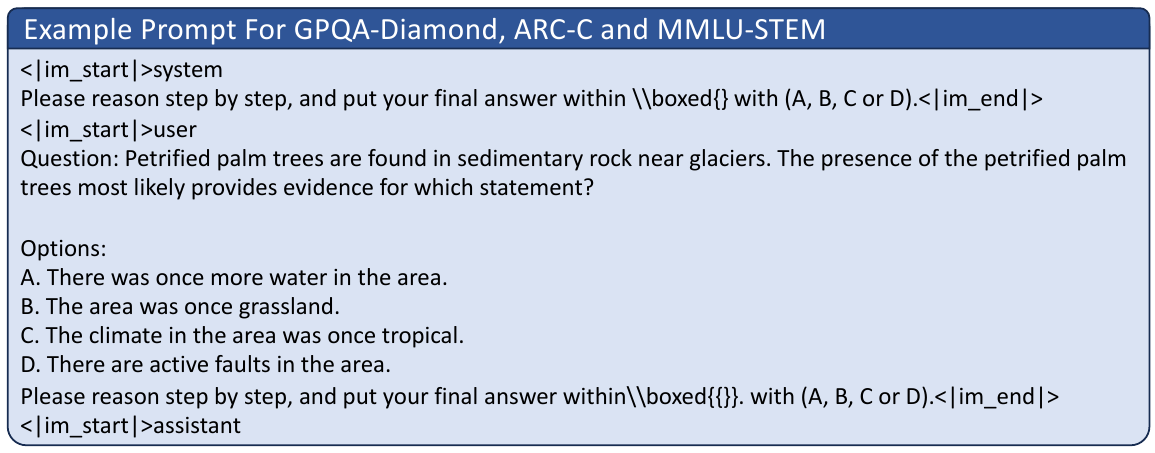}
\vspace{-7mm}
\caption{ 
Example Prompt for GPQA-Diamond, ARC-C and MMLU-STEM
 in LEPO.
}
\label{fig: prompt-fig2}
\vspace{-3mm}
\end{figure*}
We provide the specific prompt templates used in our experiments. 
As shown in Figure \ref{fig: prompt-fig1}, for mathematical reasoning tasks, the system instruction explicitly requires the model to reason step-by-step and enclose the final answer within \verb|\boxed{}|.
As shown in Figure \ref{fig: prompt-fig2}, for multiple-choice benchmarks (i.e., GPQA-Diamond, ARC-C, and MMLU-STEM), the prompt is adapted to instruct the model to provide the specific option letter (A, B, C, or D) inside the box.


\subsection{Implementation Details}

For all experiments, we utilize the TRL framework \citep{vonwerra2022trl} as the training foundation. 
We employ the AdamW optimizer with a cosine learning rate scheduler, setting the warmup ratio to $0.03$ and weight decay to $0.1$. 
The optimizer hyperparameters are set to $\beta_1 = 0.9$ and $\beta_2 = 0.99$. 
Across all models, we set the learning rate to $1 \times 10^{-6}$ and train for $1$ epoch. 
The coefficient for the KL divergence penalty, $\beta$, is set to $1 \times 10^{-3}$. 
Regarding data sampling, we use a group size of $8$ for each query and a global batch size of $32$. 
For all datasets, the maximum sequence length is set to $2048$. 
We apply top-$p$ sampling with $p=0.95$ and top-$k$ sampling with $k=30$. 
The sampling temperatures are set to $1.0$ during training and $0.6$ for testing. 
All experiments are conducted with 8 Nvidia H20 GPUs.
Training takes approximately 28 hours for Qwen2.5-7B, and 20 hours for Qwen2.5-3B and Llama-3.2-3B-Instruct.

For LEPO, we conduct a grid search over $L_g \in \{32, 64, 128, 256, 2048\}$ and $\tau_g \in \{0.3, 0.4, 0.5, 0.6, 0.7\}$, utilizing the MATH500 dataset for validation. 
Specifically, $L_g = 2048$ indicates that the entire response consists of latent tokens; in this scenario, to extract the final answer, we decode the top-1 token from the new distribution as the readable output.
The optimal hyperparameters are identified as follows: for Qwen2.5-7B, $L_g=64$ and $\tau_g=0.5$; for Qwen2.5-3B, $L_g=64$ and $\tau_g=0.3$; and for the Llama3.2-3B-Instruct model, $L_g=32$ and $\tau_g=0.5$.


\section{Additional Results}

\subsection{Comparison to Deterministic Latent Reasoning with RL}


\renewcommand{\topfraction}{0.9}
\renewcommand{\dblfloatpagefraction}{0.9}

\begin{table*}[htbp]
\centering
\caption{Comparison of stochastic versus deterministic latent reasoning on Qwen2.5-7B. We evaluate the performance using Pass@1 and Pass@32 metrics. LEPO significantly outperforms the deterministic baseline, underscoring the necessity of noise injection for enabling effective RL supervision and enhancing exploration capabilities.}
\label{tab: app-det}

\renewcommand{\arraystretch}{0.95} 

\setlength{\tabcolsep}{1.8pt} 
\vspace{-3mm}
\resizebox{\textwidth}{!}{%
\begin{tabular}{l cc cc cc cc cc cc >{\columncolor{sball}}c >{\columncolor{sball}}c}
\toprule[1.5pt]
\multirow{2}{*}{\textbf{Method}} & \multicolumn{2}{c}{GSM8K} & \multicolumn{2}{c}{MATH500} & \multicolumn{2}{c}{Minvervamath} & \multicolumn{2}{c}{AIME2024} & \multicolumn{2}{c}{AIME2025} & \multicolumn{2}{c}{AMC23} & \multicolumn{2}{c}{\textbf{Average}} \\
\cmidrule(r){2-3} \cmidrule(lr){4-5} \cmidrule(lr){6-7} \cmidrule(lr){8-9} \cmidrule(lr){10-11} \cmidrule(lr){12-13} \cmidrule(l){14-15}
 & Pass@1 & Pass@32 & Pass@1 & Pass@32 & Pass@1 & Pass@32 & Pass@1 & Pass@32 & Pass@1 & Pass@32 & Pass@1 & Pass@32 & \textbf{Pass@1} & \textbf{Pass@32} \\
\midrule

\multicolumn{15}{l}{\cellcolor{rowgray}\textbf{Qwen2.5-7B}} \\
\midrule
\quad Deterministic Latent & 88.75 & 94.62 & 73.84 & 91.80 & 32.49 & 51.10 & 8.04 & 26.67 & 6.54 & 23.33 & 47.27 & 85.00 & 42.82 & 62.09 \\
\quad \textbf{LEPO (Ours)} & \textbf{90.28} & \textbf{98.03} & \textbf{75.53} & \textbf{93.80} & \textbf{34.06} & \textbf{59.56} & \textbf{11.04} & \textbf{36.67} & \textbf{8.33} & \textbf{36.67} & \textbf{52.58} & \textbf{92.50} & \textbf{45.30} & \textbf{69.54} \\

\bottomrule[1.5pt]
\end{tabular}%
}
\vspace{-1mm}
\end{table*}
In addition, we evaluated stochastic versus deterministic latent reasoning with RL. 
Since deterministic approaches lack randomness in the reasoning process, they cannot produce the valid supervision signals required for direct guidance, limiting optimization to the final answer. 
As evidenced by Table \ref{tab: app-det}, LEPO achieves superior performance in both pass@1 and pass@32 compared to deterministic latent reasoning.
This underscores the necessity of obtaining stochastic trajectories via noise injection, which serves the dual purpose of enabling RL-based supervision for latent reasoning and enhancing the model's exploration capabilities.

\subsection{Comparison to SFT-based Latent Reasoning Methods}

\begin{table}[htbp]
\centering
\caption{Comparison with SFT-based latent reasoning methods (Coconut and CODI) on GSM8K and MATH500. All models utilize Llama-3.2-3B-Instruct as the backbone and are trained on the MATH dataset.}
\label{tab: app-sft}
\vspace{-3mm}
\resizebox{0.4\textwidth}{!}{%
\begin{tabular}{@{\hspace{2mm}}l cc}
\toprule[1.5pt]
\textbf{Method} & \textbf{GSM8K} & \textbf{MATH500} \\
\midrule

\multicolumn{3}{l}{\cellcolor{rowgray}\textbf{Llama3.2-3B}} \\
\midrule
\quad Coconut & 33.51 & - \\
\quad CODI & 69.98 & 37.80 \\
\quad LEPO (Ours) & \textbf{76.57} & \textbf{46.80} \\

\bottomrule[1.5pt]
\end{tabular}%
}
\vspace{-3mm}
\end{table}
In addition to comparing against strong RL baselines such as GRPO and HRPO, we also evaluated SFT-based latent reasoning methods, including Coconut \citep{coconut} and CODI \citep{shen2025codi}. 
Since the DAPO-MATH-17k dataset used in our main experiments lacks Chain-of-Thought (CoT) traces, we standardized the training on the MATH \citep{hendrycks2021measuring} dataset and evaluated performance on MATH500 and GSM8K. 
For all comparisons, we utilized Llama-3.2-3B-Instruct as the backbone model. 
As shown in Table \ref{tab: app-sft}, LEPO significantly outperforms both Coconut and CODI. 
This demonstrates the effectiveness of applying reinforcement learning to supervise latent reasoning processes, exhibiting its consistent advantages over prior SFT-based latent methods.

\subsection{Analysis on More Base Models}
\begin{figure*}[t]
\centering
\includegraphics[width=0.95\textwidth]{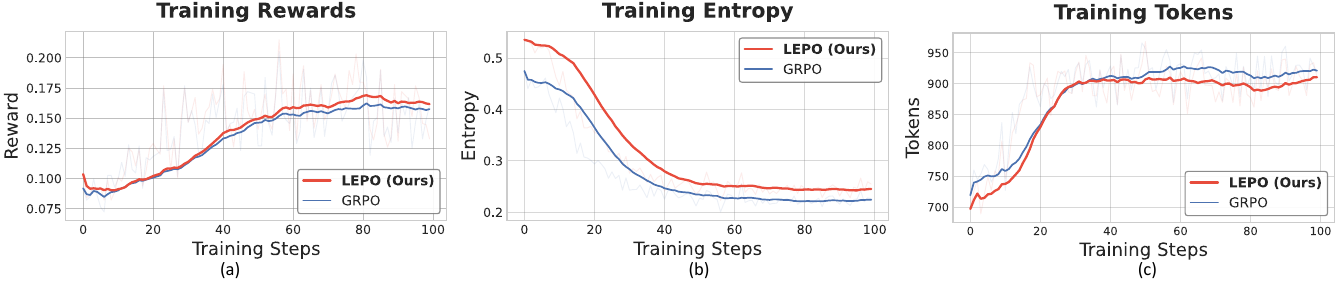}
\vspace{-3mm}
\caption{
Training analysis of LEPO versus GRPO on Qwen2.5-3B. (a) Evolution of training rewards, where LEPO outperforms GRPO. (b) Policy entropy trends; despite the initial drop and subsequent rise, LEPO maintains consistently higher entropy than GRPO. (c) Evolution of generated token counts, demonstrating the superior token efficiency of LEPO compared to the baseline.
}
\label{fig: app-qwen3b}
\vspace{-5mm}
\end{figure*}
\begin{figure*}[t]
\centering
\includegraphics[width=0.95\textwidth]{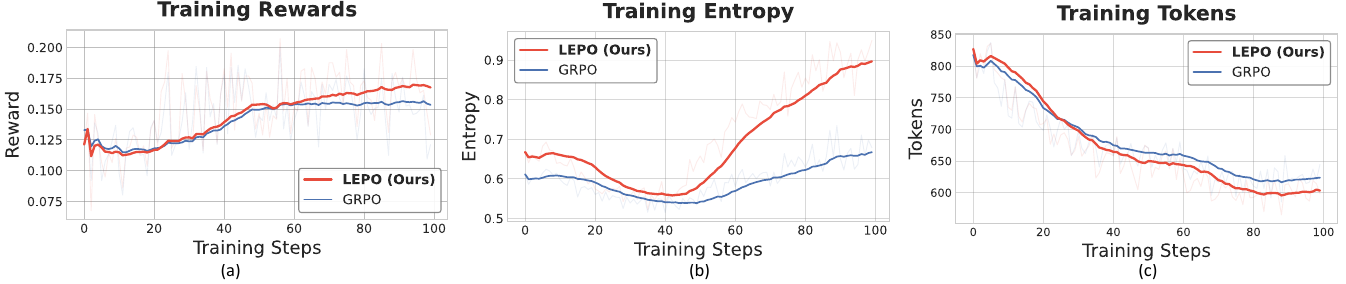}
\vspace{-3mm}
\caption{
Training analysis of LEPO versus GRPO on Llama-3.2-3B-Instruct. (a) Evolution of training rewards, where LEPO outperforms GRPO. (b) Policy entropy trends; despite the initial drop and subsequent rise, LEPO maintains consistently higher entropy than GRPO. (c) Evolution of generated token counts, demonstrating the superior token efficiency of LEPO compared to the baseline.
}
\label{fig: app-llama3b}
\vspace{-5mm}
\end{figure*}
In addition to the analysis of Qwen2.5-7B presented in Figures \ref{fig: entropy-fig}, \ref{fig: tokens-fig} and \ref{fig: ablation-fig}, we further investigate the training dynamics, entropy, and token efficiency of Qwen2.5-3B and Llama-3.2-3B-Instruct.

\textbf{Training Dynamics.}
As illustrated in Figures \ref{fig: app-qwen3b}a and \ref{fig: app-llama3b}a, we compare the training dynamics of LEPO and GRPO on the Qwen2.5-3B and Llama-3.2-3B-Instruct models. The results indicate that while both LEPO and GRPO effectively improve model performance, LEPO achieves superior final convergence.

\textbf{Entropy.}
As shown in Figure \ref{fig: app-qwen3b}b, we observe that the training entropy for Qwen2.5-3B gradually decreases, with LEPO consistently maintaining higher entropy than GRPO. Conversely, as depicted in Figure \ref{fig: app-llama3b}b, the entropy for Llama-3.2-3B-Instruct initially decreases before rising; notably, LEPO's entropy remains consistently higher than that of GRPO throughout. This demonstrates that stochastic latent reasoning possesses greater exploratory capability compared to discrete representations and maintains this heightened exploration during the training process.

\textbf{Token Efficiency.}
As illustrated in Figure \ref{fig: app-qwen3b}c, while the reasoning length of Qwen2.5-3B gradually increases during training, LEPO utilizes continuous representations to reduce the number of tokens compared to GRPO. Similarly, in Figure \ref{fig: app-llama3b}c, although the token count for Llama-3.2-3B-Instruct gradually decreases, LEPO consistently utilizes fewer tokens. These results highlight the superiority of LEPO over GRPO in terms of token efficiency.

\section{The Use of LLMs}
During the drafting phase of this work, we made use of LLMs to assist with text composition and editing. These models served primarily to enhance the flow of the narrative, adjust phrasing, and ensure grammatical correctness. Crucially, the intellectual contributions—including all hypotheses, experimental setups, and derived insights—originate solely from the authors, who retain complete accountability for the manuscript's integrity.

\end{document}